\documentclass[conference]{IEEEtran}
\IEEEoverridecommandlockouts
\usepackage{cite}
\usepackage{amsmath,amssymb,amsfonts}
\usepackage{algorithmic}
\usepackage{graphicx}
\usepackage{textcomp}
\usepackage{xcolor}
\usepackage{orcidlink} 
\usepackage{caption}
\usepackage{subcaption}
\usepackage{multirow}
\usepackage{comment}
\def\BibTeX{{\rm B\kern-.05em{\sc i\kern-.025em b}\kern-.08em
    T\kern-.1667em\lower.7ex\hbox{E}\kern-.125emX}}

\begin{document}

\title{Operationalization of Machine Learning with Serverless Architecture: An Industrial Implementation for Harmonized System Code Prediction
}

\author{
\IEEEauthorblockN{Sai Vineeth Kandappareddigari\orcidlink{0000-0001-9734-7358}}
\IEEEauthorblockA{\textit{Governance} \\
Schneider Electric U.S.A\\
Boston, USA\\
vineeth.kandappareddigari@se.com}
\and
\IEEEauthorblockN{Santhoshkumar Jagadish\orcidlink{0000-0003-2084-8124}}
\IEEEauthorblockA{\textit{Governance} \\
\textit Schneider Electric U.S.A\\
Boston, USA\\
santhoshkumar.jagadish@se.com}
\and
\IEEEauthorblockN{Gauri Verma}
\IEEEauthorblockA{\textit{Governance} \\
\textit Schneider Electric U.S.A\\
Boston, USA\\
gauri.verma@se.com}
\and
\IEEEauthorblockN{Ilhuicamina Contreras}
\IEEEauthorblockA{\textit{Tax and Customs} \\
\textit Schneider Electric Mexico\\
Mexico city, Mexico\\
Ilhuicamina.contreras@se.com}
\and
\IEEEauthorblockN{Christopher Dignam}
\IEEEauthorblockA{\textit{Governance}\\Schneider Electric\\Paris, France\\
christopher.dignam@se.com}
\and
\IEEEauthorblockN{Anmol Srivastava}
\IEEEauthorblockA{\textit{Governance}\\Schneider Electric U.S.A\\Boston, USA\\
srivastava.anm@northeastern.edu}
\and
\IEEEauthorblockN{Benjamin Demers}
\IEEEauthorblockA{\textit{Governance}\\Schneider Electric U.S.A\\Boston, USA\\
ben.n.demers@gmail.com}
}

\maketitle
\begin{abstract}
Industrializing machine learning operations (MLOps) requires architectures that deliver scalability, automation, and operational reliability. This paper introduces a serverless MLOps framework that orchestrates the entire model life-cycle (data ingestion, training, deployment, monitoring, and retraining using event-driven pipelines and managed services). The architecture is model-agnostic, supporting diverse inference patterns through standardized interfaces, enabling rapid adaptation without infrastructure overhead. To demonstrate its practical applicability, we present an industrial implementation for Harmonized System (HS) code prediction, a compliance-critical classification task in which short, unstructured product descriptions must be mapped to standardized codes used by customs authorities for global trade. Frequent updates and ambiguous descriptions make accurate classification challenging, and errors can lead to shipment delays and financial losses. Our solution employs a custom text embedding encoder and multiple deep learning architectures, with the best-performing model (Text CNN) achieving 98 percent accuracy on ground truth data. Beyond accuracy, the pipeline ensures reproducibility, auditability, and SLA adherence under variable loads, leveraging auto-scaling. A distinguishing feature of the framework is automated A/B testing, enabling dynamic model selection and safe promotion in production environments. One key consideration in model choice is cost-efficiency; while transformer-based models may achieve similar accuracy, their long-term operational costs for training and maintenance are significantly higher. While generative AI offers flexibility for exploratory tasks, deterministic classification with predictable latency and explainability is prioritized here over generalized models. Importantly, the architecture remains extensible to integrate transformer variants and LLM-based inference when appropriate. The first section of the paper introduces the deep learning architectures used, along with simulations and model comparison results. The second section discusses the industrialization of MLOps through a serverless architecture, demonstrating the feasibility of automated retraining, prediction, and validation of HS codes. This work provides a replicable blueprint for the operationalization of machine learning using serverless architecture, enabling enterprises to scale confidently while optimizing both performance and economics.
\end{abstract}



\maketitle
\section{Introduction}
\label{sec:introduction}
\subsection{Business Problem}
\IEEEPARstart{C}{ustoms} and the trade compliance organization is responsible for monitoring products shipped internationally. Any product transported across national borders requires the correct assignment of HS codes. The Harmonized Commodity Description and Coding System, commonly referred to as the Harmonized System of Tariff Nomenclature, is a globally standardized system that provides names and numerical codes for the classification of traded products. Developed by the World Customs Organization (WCO), this system includes approximately 5,000 product categories, each identified by a six-digit code.These codes are structured hierarchically: the first two digits represent the chapter, the next two denote the heading, and the final two specify the subheading. The classification is further supported by implementing regulations and explanatory notes \cite{b26}. The Harmonized System enables economic operators, customs officials, and legislators worldwide to identify and classify the same product using a universally recognized numeric code.

The assignment of HS codes is time consuming and requires significant diligence to ensure accuracy and minimize human error \cite{b27}. This process becomes particularly challenging because exported goods often include both raw materials destined for manufacturing hubs and finished products with commercial value. As a result, the volume of products that require HS code classification is substantial, and assigning the correct code to a finished product can be complex. Furthermore, regional policy changes can cause countries to lose track of code assignments, leading to discrepancies in the use of HS codes across borders. These inconsistencies can delay international shipments and disrupt logistics operations. In response to these challenges, customs authorities are increasingly exploring the potential of machine learning to automate, audit, and resolve issues related to the classification of HS codes \cite{b1}.

In order to prevent customs code misalignment between countries, there is an imminent need to harmonize the codes across regions. A Machine Learning (ML) approach has been adopted to harmonize codes in different countries and strengthen code assignments \cite{b7}, \cite{b11}, \cite{b28}. The ML model will be utilized as part of the assignment process to validate correct code assignments before products are shipped. Harmonized codes significantly reduce mismatches between products and their corresponding HS codes, thereby minimizing delays in international trade. This approach not only improves consistency but also facilitates smoother and more efficient export processes.

To address these challenges at scale, the solution incorporates Machine Learning Operations (MLOps) principles to ensure that models remain accurate and reliable over time. MLOps enables automated retraining and monitoring as product descriptions and HS code definitions evolve, reducing manual intervention and minimizing classification errors. Additionally, the framework integrates A/B testing to compare multiple models under real-world conditions before full deployment. This approach ensures that the most effective model is promoted based on performance metrics such as accuracy, latency, and cost-efficiency. By combining automation with dynamic model selection, customs organizations can maintain consistent compliance while adapting quickly to regulatory changes and fluctuating trade volumes

This research paper is organized into eight sections. Sections 1 through 4 cover the literature review, data preparation, evaluation of suitable algorithms using A/B testing, and model training to automate HS code classification. Sections 5 and 6 present the evaluation strategy, performance metrics, and final results. Section 7 focuses on the industrialization of MLOps through a serverless architecture implemented on AWS. Finally, Section 8 provides the conclusion and outlines directions for future research.

\section{RELATED WORK}
Most of the machine learning work seen in HS codes is implemented using combined models. One such case includes textual and visual information about products with limited data\cite{b6},\cite{b29}, \cite{b8}. The fields used for modeling the use case of customs code classification are product descriptions (which may or may not contain technical specifications). However, most of the work is done on algorithms that support textual information. Additionally, models have been built on top of large datasets, which are implemented using GPU resources on cloud infrastructure.

In \cite{b30}, the work presented follows a radical approach using the HS Code ontology. The work in this paper varies from the approach mentioned above. In this paper, the concepts of the HS Code ontology are not used. Only text descriptions associated with various products are used. This is because the training data contains products that are closely aligned and belong to similar chapters in the HS code. However, this approach is not well suited for use cases where the products have hierarchical HS codes. For example, products may have HS codes A and B together. However, HS Code A is a subset of HS Code B. Although ontology adds knowledge to the construction of the classification model, this approach is not used in this paper. 

In \cite{b29} for the classification of the HS code, a characteristic model is presented using the description of the goods. As one of the comparative works on text classification using supervised machine learning methods resulted in SVM as a good performing model \cite{b17}, our baseline model implementation included SVM as a possible approach.

As seen in \cite{b19}, to overcome the limitations of linear algorithms, CNN's were explored as a possible solution. CNN's are designed to exploit the "spatial correlation" in data and work best on images and speech. Therefore, compared to the work done in \cite{b6}, \cite{b3}, the implementation of LSTM, DNN, and Text-CNN in this paper shows a better grasp of memorizing textual patterns, as they retain long-term dependencies between word sequences while handling the nonlinearity of the network. Since these models require more hyperparameters, automated hyperparameter tuning using a Bayesian approach has been incorporated into the models to create design choices, as seen in \cite{b18}.

In addition to the current implementation in the paper, the Neural Architecture Search (NAS) technique has gained attention for automatically designing deep learning networks. NAS has shown promise in optimizing architectures for specific tasks, offering more efficient and effective models\cite{b32}\cite{b33}. 

Furthermore, the concept of Zero NAS, which aims to minimize the computational resources required for NAS, has emerged as a cutting-edge approach in the field of deep learning\cite{b31}. In addition, the use of DSNet++ will help speed up inference and reduce deployment complexity in our infrastructure\cite{b33}. The continued exploration and refinement of these techniques promise to unlock new frontiers in text classification, empowering systems to comprehend and interpret textual data with unprecedented depth and precision. Eventually, when the HS Code ML architecture becomes complicated, adapting these approaches will be considered to reduce time and computational complexity.

Although the existing literature \cite{b3},\cite{b6}, \cite{b8}, \cite{b28}, \cite{b29} has focused on experiments with various classification methods and feature encoding techniques, the incorporation of NAS and the latest generative models in the context of text classification presents a new frontier, promising significant advancements in the field.

In addition to the aforementioned advances, the integration of the NAS (Neural Architecture Search) technique could further enhance the model architecture by automating the process of finding the optimal network configurations for LSTM and Text-CNN, potentially leading to improved performance and reduced complexity \cite{b35}.

As seen in \cite{b3}, the work presented for HS classification focuses on experiments with three methods: Libshorttext, text categorization, and topic modeling, followed by the implementation of the SVM and KNN models.
Similarly, in \cite{b29}, Random Forest is implemented in addition to the baseline paper, which presented multinomial NB, KNN, and decision trees for the same use case.
In \cite{b29}, one hot encoding is applied to encode nominal and categorical variables, combined with TF-IDF vectorization. However, this does not relate to the work in this paper, as categorical features are not incorporated into the models.

After carefully reviewing the concepts used in the papers mentioned above, neural network implementation has been chosen as it exhibits temporal behavior and captures sequential data, making it a practical approach compared to the classical ones \cite{b16}, \cite{b12}, \cite{b19}, \cite{b20}.


\section{Overview}
\subsection{About the Data}
Effective machine learning for HS code harmonization requires high-quality training data from trusted sources. The dataset includes product descriptions and technical specifications, along with previously assigned HS codes for shipped items. HS codes serve as the target (dependent) variable, while product information acts as the input features (independent variables). Prior to model training, internal export control experts validated the code assignments. Each data point was then categorized into assurance levels 1 through 4, based on data quality.

Assurance levels 1 and 2 indicate that the assigned HS codes are likely inaccurate and require further expert review. In contrast, levels 3 and 4 represent high-confidence labels and are used for model training. Efforts were made to improve the quality of data classified under levels 1 and 2 through additional validation. This process ensures that the model is trained on stable, verified data. Moreover, incorporating newly validated data over time further enhances the overall quality of the training dataset.

\subsection{Data Preparation}
\label{DataPreparation}

Data labeled with assurance levels 3 and 4 were reviewed to ensure higher stability compared to lower levels. This resulted in a more reliable dataset for model training. Consequently, the models achieved high confidence and accuracy without compromising flexibility or generalization.

\begin{figure}[h]
    \centering
    \includegraphics[scale=1]{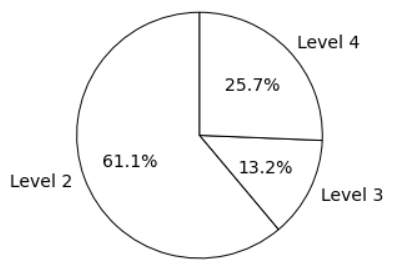}
    \caption{Distribution of assurance levels in the training data}
    \label{fig:alevels}
\end{figure}

\textbf{Figure \ref{fig:alevels}} illustrates the distribution of assurance levels in the training data after filtering out assurance level 1. 

The selection of text features is a critical component of effective information retrieval. In this study, relevant features were carefully extracted from the dataset to accurately represent and distinguish HS codes.

\textbf{Table \ref{tab:features}} explains the predictor variables (text features) that are used to predict the response variable (HS code). These text descriptions serve as input features for ML models. 

\begin{figure}[h]
    \centering
    \includegraphics[scale=0.2]{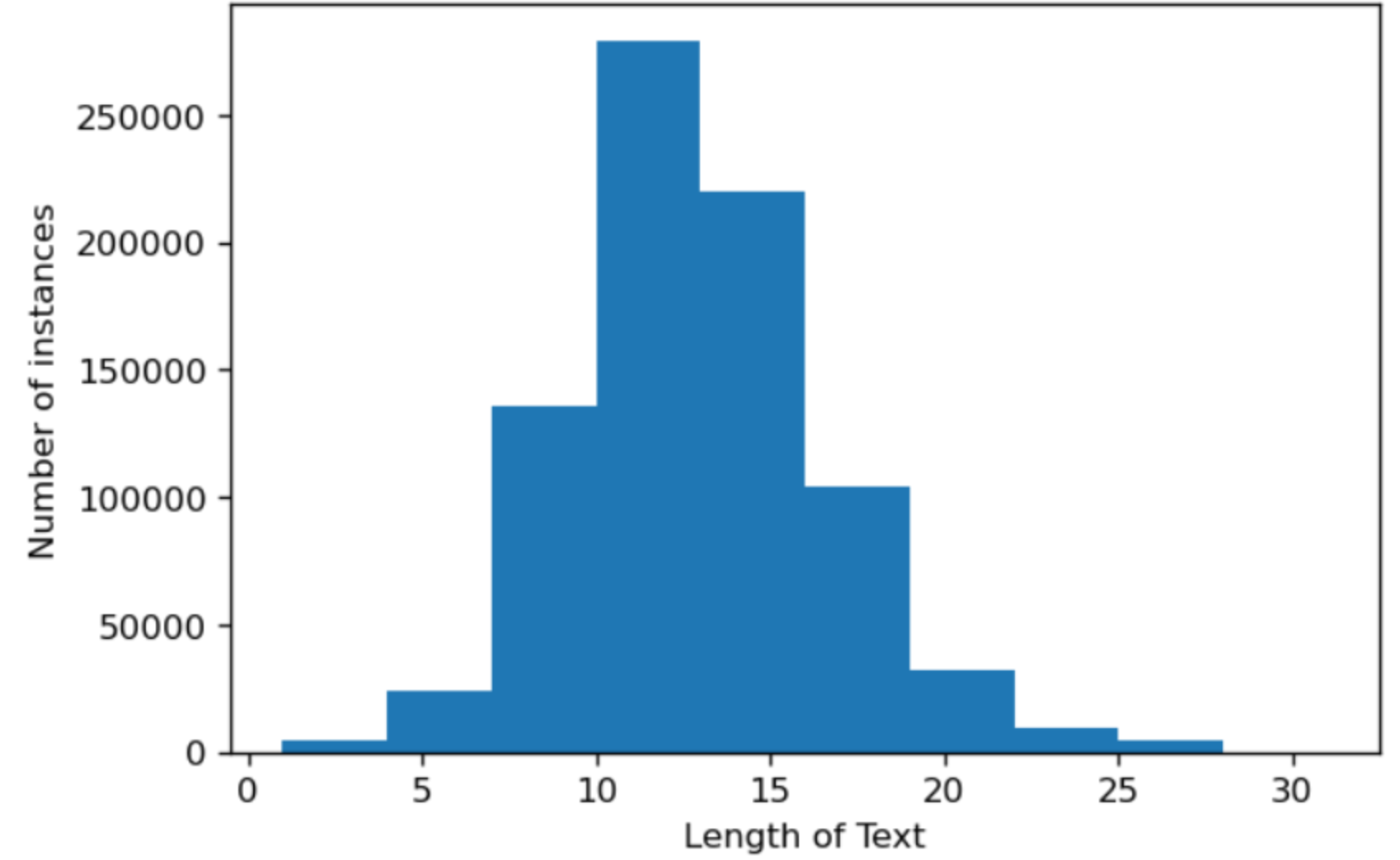}
    \caption{Text length (number of characters) for combined description in training data}
    \label{fig:comdesc}
    
    \captionsetup{justification=centering,margin=2cm}
\end{figure}

\begin{table}
    \centering
    \begin{tabular}{|l|p{50mm}|}
    \hline
    FEATURES & DESCRIPTION \\
    \hline
     Short Description &  Technical description of each item or product along with information about product hierarchy and product name\\
     \hline
     Medium Description & Description about product family that contains similar products, and also includes product dimensions and technical specifications \\
     \hline
     ETIM & International classification standard for technical products \\
     \hline
    \end{tabular}
    \caption{Features used in various models}
    \label{tab:features}
\end{table}
The average text length (characters per data point) for the medium description is 63.9, while for the short description, it is 15.2. The average text length in the combined description field is 12.4, as shown in \textbf{Figure \ref{fig:comdesc}}.

\subsection{Class Imbalance}
The data set consists of products that are shipped to different regions. Certain products are shipped in larger quantities, whereas some are shipped in smaller quantities. The difference in the volume of products shipped leads to an unequal representation of HS codes (classes) in the data set. These classes cause bias in the predictions, as the distribution of products across the known classes is skewed. Imbalanced classification is a challenge for predictive modeling, as models are designed with the assumption that all classes are equally represented. When classes are not well represented, this results in poor model performance. To overcome this, specialized techniques such as up-sampling have been implemented. 


Before up-sampling, the row count was 815,264, and after up-sampling, the row count increased to 818,048. \textbf{Figure 3} represents the class imbalance in the training data. 

Various up-sampling techniques, including SMOTE, borderline SMOTE, and random up-sampling, were considered. Stratified sampling involves duplicating examples from the minority class at random to augment its representation, thereby preserving the original data points. In contrast, SMOTE generates synthetic samples for the minority class by interpolating existing samples from the minority class, creating new artificial data that reflect the characteristics of the minority class, thus offering a more diverse representation.

In customs code classification, stratified up-sampling was preferred due to its ability to maintain the original data and mitigate potential overfitting caused by synthetic samples, which was effectively observed during experiments with SMOTE. Moreover, stratified up-sampling is simpler and more transparent in its approach and is easy to evaluate. Upon evaluating the implementation of these strategies and considering the nature of class imbalance, stratified up-sampling emerged as the most suitable approach to address this problem.
 
Classes that have less than 1 percent representation in the data set were considered the minority class. The mean and median representations of the minority class were calculated. After this, each minority class was up-sampled with a stratified approach using the mean and median ratios. This would introduce a balance within all classes without introducing unwanted bias.

\begin{figure}[h]
    \centering
    \includegraphics[scale=0.5]{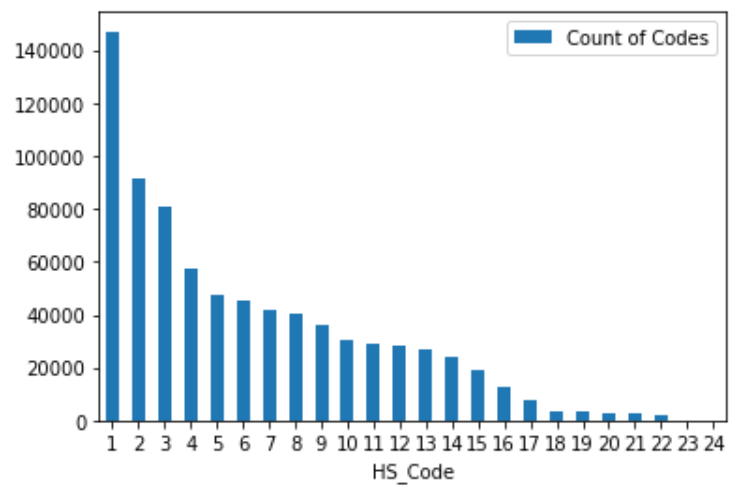}
    \caption{Class Imbalance as seen on training data}
    \label{fig:codes}
    \captionsetup{justification=centering}
\end{figure}

\subsection{Approaches}
To build the models, data belonging to the EU and US regions were selected. With respect to the modeling framework, multiple simulations were performed based on the following train-test combinations:

\begin{itemize}
\item Model without up-sampling: Trained on all codes (95\% of data for each code was added to the training set), and tested on all codes (5\% of data for each code was added to the test set) 
\item Model with up-sampling: Training of all HS codes (95\% of data for each code was added to the training set) and testing of all HS codes (5\% of data for each code was added to the test set). 
\end{itemize}

The dataset is divided using a stratified approach to ensure that each HS code has a proportional representation in training and testing samples.

\subsection{Feature Engineering}

We studied multiple features within our dataset, such as product descriptions, taxonomy, and various ETIM standard-based attributes. Due to multiple HS codes belonging to the same taxonomy family, the taxonomy itself does not serve as a unique identifier for an HS code. Following a thorough analysis of the dataset and its distinct feature representations, we opted to use short and medium descriptions as our training features. Both descriptions provide text-based information about the products. The following steps were implemented to prepare the dataset:
\begin{itemize}
\item Remove whitespace, symbols, and stop words from the text, and convert the text to lowercase.
\item Create tokens for the text descriptions using the Tokenizer function
\item Set vocabulary size: define how many unique words are to be included in the vocabulary.
\end{itemize}

Word2Vec and GloVe were used to create embeddings\cite{b34}. However, the utilization of pre-trained embeddings did not yield the desired results, given the specificity of the engineering practices and global trade descriptions utilized in our study. Consequently, we opted to incorporate custom embeddings as an integral component of the training layers, which yielded the desired results.

\subsection{Hyper-parameter Tuning}
Hyperparameters are crucial components for the effective training of ML models. They help tailor the behavior of the model to improve training time and the learning process. They also help control model convergence and improve model performance while reducing infrastructure requirements and computation time. The hyperparameters tuned for the LSTM and DNN models are listed in \textbf{Table \ref{tab:hyp-dnn-lstm}}. In addition, the Text-CNN hyperparameters are listed in \textbf{Table \ref{tab:hyp-TextCNN}}.

\begin{itemize}
\item{\verb|Dropout:|} Percent of edges that are randomly dropped from hidden neural units. In order to avoid overfitting, dropout layers are used between subsequent layers, and the dropout rate varies from $10\%$ to $75\%$.
\item{\verb|neuronPct:|} The percent of neurons in the range 0-5000 (user-defined range) that should be used in the entire network 
\item{\verb|neuronShrink:|} Neural networks usually start with more neurons in the first hidden layer and then decrease the number of neurons in subsequent layers. The addition of more layers is stopped when there are no more neurons available(with the count specified by neuronPct).
\item{\verb|initialNeurons:|} Count of neurons in the first layer of the neural network. Its range varies from 11 to 174, which represents the mode of document length in the dataset before and after the removal of stop words, respectively.
\item{\verb|dim embedding:|} The embedding dimension should be less than the vocabulary size extracted from short and medium descriptions. In addition, the dimension should be able to address every latent relationship available in these descriptions.
\item{\verb|n layer:|} Total number of layers in the neural network.
\end{itemize}
Additional Text-CNN hyperparameters are listed below:

\begin{itemize}
\item{\verb|Filter Size:|} Filters define the size of the output vector for each convolution
\item{\verb|Kernel Size:|} It helps specify the length of the one-dimensional convolution window, that is, the occurrence of words together(1-gram, bi-gram, or n-gram approach)
\end{itemize}
A well-thought-out range for each of these hyperparameters was passed to the Bayesian Optimizer.

\section{Study Methodology}
\subsection{Classical Approach}
The text classification approach using the Support Vector Classifier (SVC) was initially considered for modeling. SVC is based on bag-of-words representations that are free of any order. After training with multiple combinations of features, the accuracy obtained was 85\%. Taking into account the results with a confidence greater than 80\%, 8.1\% of the samples were incorrectly predicted. However, for outcomes with a confidence less than 80\%, 11.15\% of the samples were incorrectly predicted. The time taken for hyperparameter tuning and training was significantly longer. For these reasons, neural network approaches were preferred. Other possibilities, such as using word embeddings as input to SVC, were not explored.\cite{b17}

\subsection{Long-Short Term Memory (LSTM)}
For all neural network models, the word embedding is used as the first layer instead of pre-trained embeddings. Since the descriptions in the HS codes are specific to the products to be shipped and do not follow the general sentence structure, pre-trained embeddings are not useful\cite{b13}. As seen in \cite{b20}, the LSTM is a sequence architecture. Unlike standard Neural Networks, LSTMs have feedback connections that help to process entire data sequences\cite{b4} \cite{b15}. LSTMs are proven to be well suited for classification and making predictions based on sequence datasets. They help deal with the problem of vanishing gradients that can be encountered when using traditional RNN. Since the training data set has text fields that give relevant information about the HS codes, LSTM models are used.

\begin{figure*}[h]
    \centering
    \begin{subfigure}[b]{\textwidth}
        \centering
        \includegraphics[width=\linewidth]{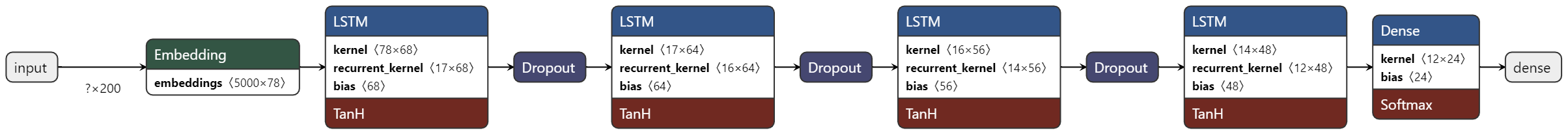}
    \end{subfigure}
    \vfill
    \begin{subfigure}[b]{\textwidth}
        \centering
        \includegraphics[width=\linewidth]{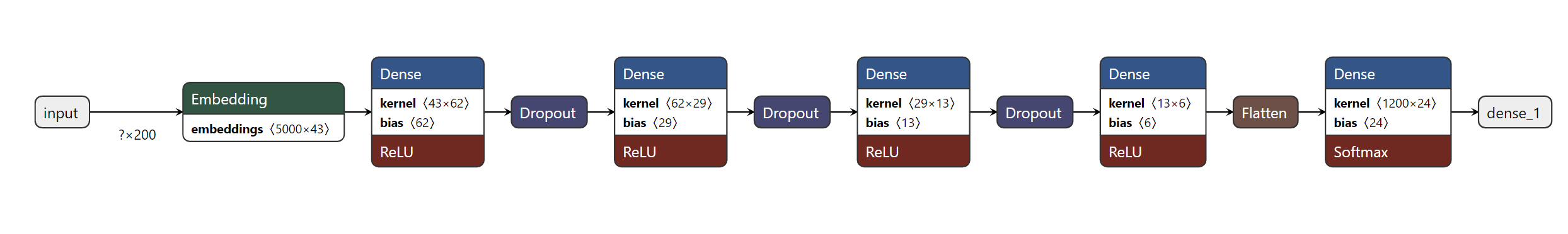}
    \end{subfigure}
    \begin{subfigure}[b]{\textwidth}
        \centering
        \includegraphics[width=\linewidth]{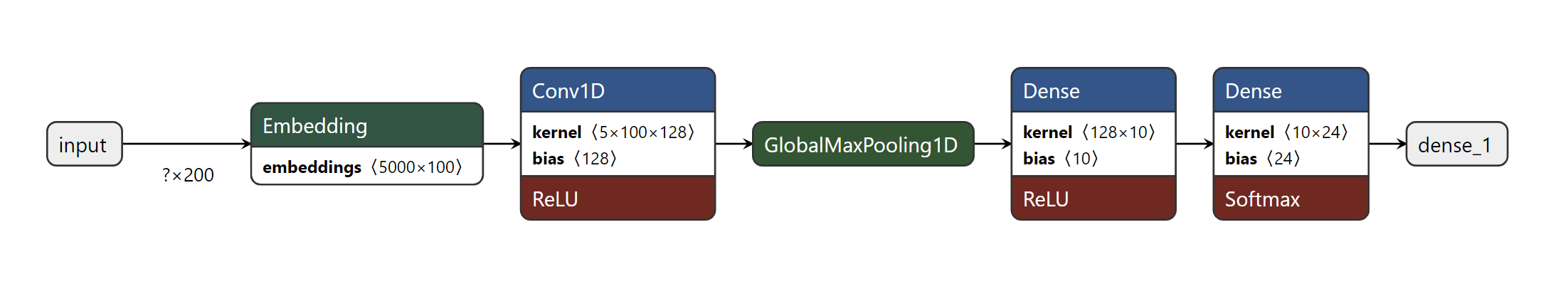}
    \end{subfigure}
\caption{Top to Bottom: Architecture Diagram for LSTM, DNN and Text-CNN Model (Specifications in the diagram are the same as those used in the experimentation) }
\label{fig:architecture-diagrams}
\end{figure*}

As seen in \textbf{Figure \ref{fig:architecture-diagrams}}, LSTM Architecture is implemented. Using the Bayesian Optimizer, the described set of hyperparameters was adjusted for model training \cite{b18}, as illustrated in \textbf{Table \ref{tab:hyp-dnn-lstm}}.

\subsection{Deep Neural Network (DNN)}
DNN is a collection of multiple perceptrons or neurons that form a single layer of neurons \cite{b2}. These layers can be stacked together in a sequential manner, where the current neural layer receives input from the previous layer and performs mathematical computations using activation functions. Since the descriptions of the products represent the product's specifications and have no contextual significance, DNN was chosen for experimentation, in addition to LSTM and Text-CNN. \textbf{Figure \ref{fig:architecture-diagrams}} illustrates the architecture of DNN.\cite{b10} The values of the hyperparameters are determined by the Bayesian optimizer. It is illustrated in \textbf{Table \ref{tab:hyp-dnn-lstm}} for the model trained on data before and after up-sampling. 

\subsection{Text-CNN (Convolutional Neural Network)}
Recent studies show that CNN performs well in NLP problems. RNNs are well versed in classification tasks if determined by a long-range semantic dependency instead of local key phrases, whereas CNNs perform well at extracting local and position-invariant features. The other benefit of CNNs is that they are fast compared to RNNs when implemented using a GPU. This is because convolutions are an integral part of GPU matrices. Based on these characteristics, Text-CNN was used for HS Codes Classification. In addition, descriptions are short, and key phrases in descriptions can strengthen the CNN task. Text CNNs are one-dimensional that capture the association between adjacent words\cite{b16} This groundbreaking approach of applying CNN to text classification first appeared in 2014 \cite{b5}.

\textbf{Figure \ref{fig:architecture-diagrams}} illustrates the Text-CNN architecture that has been implemented. The approach is to embed the words of a sentence and pass them to parallel convolutions with different kernel sizes to mimic the n-gram approach\cite{b4}.This is followed by the activation function, max pooling, and concatenation. Concatenation is used to embed the sentence as a single vector before passing it on to a fully connected layer, followed by a softmax function. The softmax function generates the probability values for the HS codes. Refer to  \textbf{Table \ref{tab:hyp-TextCNN}} for the range of hyperparameters and the values that are finalized for the model.

The operations involved in the model architecture are as follows:
\begin{itemize}
\item{\verb|Embedding|}: Embed a batch of text with shape (N, L) to a shape (N, L, D), where N is the batch size, L is the maximum length of the batch, and D is the embedding dimension
\item{\verb|Convolutions|}: Run parallel convolutions across the embedded words with kernel sizes of 3, 4, and 5 to mimic trigrams, four grams, and five grams. This results in outputs of shape (N, L - k + 1, D) per convolution, where k is the kernel size.
\item{\verb|activation|}: ReLU activation function is applied for each convolution operation.
\item{\verb|Pooling|}: Run parallel max-pooling operations on the activated convolutions with a window size of (L - k + 1). This results in 1 value per channel, that is, a shape of (N, 1, D) per pooling.
\item{\verb|Concat|}: The pooling output is concatenated and squeezed to produce the shape (N, 3D) that represents a single embedding for a sentence.
\item{\verb|Fully Connected|}: The output from the pooling layer is passed through a fully connected layer with the shape (3D, 1) to yield a single output for each example in the batch. The sigmoid activation function is applied to the output of this layer.
\end{itemize}

\subsection{\textbf{Mathematical Formulation of Text-CNN}}

Let a sentence be represented as a sequence of words:
\[
S = [w_1, w_2, \dots, w_L]
\]
where \( L \) is the maximum sentence length. Each word \( w_i \) is assigned to a dense vector using an embedding matrix \( \mathbf{E} \in \mathbb{R}^{V \times D} \), where \( V \) is the size of the vocabulary and \( D \) is the embedding dimension. The embedded sentence becomes:
\[
\mathbf{X} = [\mathbf{x}_1, \mathbf{x}_2, \dots, \mathbf{x}_L] \in \mathbb{R}^{L \times D}
\]

\textbf{Convolution Layer:} A 1D convolution operation with a filter \( \mathbf{W}_k \in \mathbb{R}^{h \times D} \) (where \( h \) is the kernel size) is applied to a window of \( h \) words to produce a feature:
\[
c_i = f(\langle \mathbf{W}_k, \mathbf{X}_{i:i+h-1} \rangle + b)
\]
where \( \langle \cdot, \cdot \rangle \) denotes the Frobenius inner product, \( b \) is a bias term, and \( f \) is a non-linear activation function (ReLU).

This operation is applied across all valid windows to produce a feature map:
\[
\mathbf{c} = [c_1, c_2, \dots, c_{L - h + 1}]
\]

\textbf{Max-Pooling Layer:} A max-over-time pooling operation is applied to each feature map:
\[
\hat{c} = \max\{\mathbf{c}\}
\]

This results in one scalar per filter. If \( F \) filters are used per kernel size and \( K \) different kernel sizes are applied, the output of the pooling layer is:
\[
\mathbf{z} \in \mathbb{R}^{F \cdot K}
\]

\textbf{Fully Connected Layer:} The pooled features are passed through a fully connected layer:
\[
\mathbf{o} = \mathbf{W}_{fc} \mathbf{z} + \mathbf{b}_{fc}
\]
where \( \mathbf{W}_{fc} \in \mathbb{R}^{C \times (F \cdot K)} \), \( \mathbf{b}_{fc} \in \mathbb{R}^{C} \), and \( C \) are the number of classes (50 HS codes).

\textbf{Softmax Output:} The final output is obtained using the softmax function:
\[
\hat{y}_i = \frac{e^{o_i}}{\sum_{j=1}^{C} e^{o_j}}, \quad \text{for } i = 1, \dots, C
\]

\textbf{Loss Function:} The model is trained using the Categorical Cross-Entropy loss:
\[
\mathcal{L} = - \sum_{i=1}^{C} y_i \log(\hat{y}_i)
\]
where \( y_i \) is the one-hot encoded true label and \( \hat{y}_i \) is the predicted probability for class \( i \).

\begin{table*}[hbt!]
\centering
\begin{tabular}{|l|c|c|c|c|}
\hline
\textbf{Hyper-parameters} & \textbf{Range} & \textbf{DNN base model} & \textbf{DNN up-sampled model} & \textbf{LSTM (base \& up-sampled)}\\ \hline \textbf{Initial neurons}  & 11- 174  & 11.01  & 168.01    &      25.0                    \\ \hline
\textbf{Neuron Pct}       & 0.35 - 1.0   & 0.44   & 0.95   &        1.0              \\ \hline
\textbf{Dropout}          & 0.10 - 0.75    & 0.37   & 0.1   &       0.1                       \\ \hline
\textbf{Neuron Shrink}    & 0.25 - 0.95    & 0.31  & 0.466   &      0.95                      \\ \hline
\textbf{Embedding Dim}    & 11.0  - 87.0   & 65.74  & 43.3    &   78.0                           \\ \hline
\textbf{N\_layer}         & 1.0 - 15.0     & 8.02   & 7.89     &      3.0                     \\ \hline
\end{tabular}
\caption{Hyper-parameters for DNN and LSTM base and up-sampled models}
\label{tab:hyp-dnn-lstm}
\end{table*}


\begin{table*}
\centering
\begin{tabular}{|c|c|c|c|c|c|c|c|c|c|c|c|c|c|c|c|c|c|c|c|c|}
\hline
\multirow{1}{*}{Model} & 
\multicolumn{2}{c|}{{Accuracy}} & 
\multicolumn{6}{c|}{Precision}  & 
\multicolumn{6}{c|}{Recall}  & 
\multicolumn{6}{c|}{F1} 
\\ 
\cline{4-21}
& \multicolumn{2}{l|}{}  & \multicolumn{2}{l|}{$\ge$ 90} & \multicolumn{2}{l|}{80 - 90} & \multicolumn{2}{l|}{$\le$ 80} & \multicolumn{2}{l|}{$\ge$ 90} & \multicolumn{2}{l|}{80 - 90} & \multicolumn{2}{l|}{$\le$ 80} & \multicolumn{2}{l|}{$\ge$ 90} & \multicolumn{2}{l|}{80 - 90} & \multicolumn{2}{l|}{$\le$ 80}  \\ 
\cline{2-21}
                       & B    & U                                       & B  & U                  & B & U                        & B & U                    & B  & U                  & B & U                        & B & U                    & B  & U                  & B & U                        & B & U                     \\ 
\hline
DNN                    & 89\% & 95\%                                    & 15 & 19                 & 5 & 1                        & 3 & 3                    & 12 & 20                 & 5 & 2                        & 6 & 1                    & 14 & 19                 & 5 & 2                        & 4 & 2                     \\ 
\hline
LSTM                   & 93\% & 97.50\%                                 & 19 & 20                 & 2 & 0                        & 2 & 3                    & 16 & 22                 & 5 & 1                        & 2 & 0                    & 18 & 21                 & 3 & 1                        & 2 & 2                     \\ 
\hline
Text-CNN               & 98\% & 98\%                                    & 19 & 20                 & 2 & 1                        & 2 & 2                    & 20 & 21                 & 2 & 2                        & 1 & 0                    & 19 & 20                 & 2 & 2                        & 2 & 1 \\

\hline
\end{tabular}
\caption{Accuracy, Precision, Recall and F1 metrics for DNN, LSTM and Text-CNN on training data}
\label{tab:results-train}
\end{table*}


\begin{table*}
\centering
\begin{tabular}{|c|c|c|c|c|c|c|c|c|c|c|c|c|c|c|c|c|c|c|c|c|}
\hline
\multirow{1}{*}{Model} & 
\multicolumn{2}{c|}{{Accuracy}} & 
\multicolumn{6}{c|}{Precision}  & 
\multicolumn{6}{c|}{Recall}  & 
\multicolumn{6}{c|}{F1} 
\\ 
\cline{4-21}
& \multicolumn{2}{l|}{}  & \multicolumn{2}{l|}{$\ge$90} & \multicolumn{2}{l|}{80 - 90} & \multicolumn{2}{l|}{$\le$80} & \multicolumn{2}{l|}{$\ge$90} & \multicolumn{2}{l|}{80 - 90} & \multicolumn{2}{l|}{$\le$80} & \multicolumn{2}{l|}{$\ge$90} & \multicolumn{2}{l|}{80 - 90} & \multicolumn{2}{l|}{$\le$80}  \\ 
\cline{2-21}
                       & B    & U                                       & B  & U                  & B & U                        & B & U                    & B  & U                  & B & U                        & B & U                    & B  & U                  & B & U                        & B & U                     \\ 
\hline
DNN                    & 89\% & 95\%                                    & 15 & 19                 & 5 & 1                        & 3 & 3                    & 12 & 20                 & 5 & 2                        & 6 & 1                    & 14 & 19                 & 5 & 2                        & 4 & 2                     \\ 
\hline
LSTM                   & 92\% & 97.00\%                                 & 19 & 20                 & 2 & 0                        & 2 & 3                    & 16 & 22                 & 5 & 1                        & 2 & 0                    & 18 & 21                 & 3 & 1                        & 2 & 2                     \\ 
\hline
Text-CNN               & 98\% & 98\%                                    & 19 & 20                 & 2 & 1                        & 2 & 2                    & 20 & 21                 & 2 & 2                        & 1 & 0                    & 19 & 20                 & 2 & 2                        & 2 & 1 \\

\hline
\end{tabular}
\caption{Accuracy, Precision, Recall and F1 metrics for DNN, LSTM and Text-CNN on test data}
\label{tab:results-test}
\end{table*}

\section{EVALUATION STRATEGY and METRICS}

\subsection{Loss Function}

Neural network models learn mappings from input to output from examples, and the choice of loss function must match the framework of the specific predictive modeling problem. As our use case is a multiclass classification with a limited number of classes, we opt for categorical cross-entropy as our loss function \cite{b9}.

\subsection{Performance Metrics}
Performance metrics play a crucial role in building an optimal classifier. Thus, the appropriate selection of suitable metrics is a key step in the optimal evaluation of the results. The most widely used metric is accuracy. However, it is not a reliable metric for an imbalanced dataset. In such use cases, more importance should be given to metrics such as precision and recall. While precision describes how many of the predicted codes are truly relevant, recall helps to measure how well the model can predict the relevant codes.

\subsection{A/B Testing}
A/B testing can often be used to test and improve a set of machine learning models. It can help decide whether a particular new model works better than other existing models. Three predictive models using DNN, LSTM, and Text-CNN have been developed for the use case. In order to pick a statistically significant model, A/B testing has been implemented. To do this, a k-fold cross-validation\cite{b14} with 37 folds has been performed to obtain the performance metrics across all models for each HS code. As the results did not alter much after 37 folds, it was ensured that cross-validation adequately exposes the models to randomness within the data. Then it was followed by an ANOVA test using results from the k-fold simulation as listed in the following steps:

\begin{enumerate}

\item {\verb|Cross Validation with DNN|}: The entire data set was evaluated using a Deep Neural Network (DNN) model with K-fold cross-validation. In this experiment, 37 folds were used, resulting in 37 precision and recall values for each HS code label

\item {\verb|Metric Aggregation|}: For each of the 24 HS code labels, the mean and median of the 37 precision and recall values were calculated. This resulted in 888 precision and recall values for each label. After averaging, each of the 24 HS codes was represented by a single precision and recall value, resulting in a set of metrics per code.

\item {\verb|Repetition across Models|}: Steps 1 and 2 were repeated for LSTM and Text-CNN models

\item {\verb|Dataset for ANOVA|}: The next step involved constructing two datasets, one for precision and one for recall to perform ANOVA. Each data set contained three columns, corresponding to the average metric values produced by the DNN, LSTM, and Text-CNN models. These datasets serve as the basis for statistical comparison across models for each HS code

\item {\verb|Statistical Significance Testing|}: To assess the statistical significance of the results, the K-fold metric distributions were first transformed into approximate Gaussian distributions. A one-way ANOVA test was then conducted for each HS code to compute the F-statistic and corresponding p-value

\item {\verb|Hypothesis Testing|}: The next step involved performing hypothesis testing to determine whether the performance differences between the LSTM, Text-CNN, and DNN models are statistically significant and can be meaningfully compared.

\end{enumerate}

\textbf
\textbf Hypothesis for the A/B testing: 

\begin{itemize}
\item {\verb|Null hypothesis|}: To test if the distribution of the precision or recall between models are different
\item {\verb|Alternate hypothesis|}: To test if the distribution of precision or recall between models are the same
\end{itemize}

\textbf With the defined hypothesis, ANOVA is applied to understand which models can be compared. This test helps to reject or accept the null hypothesis. The rejection of the null hypothesis validates the statistical significance of the model recommendation. After hypothesis testing, the model with the highest average metric value was identified and recommended for each HS code.

\section{Results}
 HS code classification models (DNN, LSTM and Text-CNN) follow a custom embedding approach, and the models use the same dataset for training and testing. This ensures that the performance of the model is comparable and reliable. The performance of the model is based on accuracy, precision, and recall. The accuracy metric is used to see how well the model performs on all 24 HS codes together. Class imbalance is observed when model performance goes below 80 percent, and at that point up-sampling is applied to improve accuracy. In the next steps, precision and recall are observed for each HS code. If there is any bias between precision and recall, the F1 measure with varying beta is used to evaluate the performance of the model. A higher precision score is observed for test data compared to recall for all models, as shown in \textbf{Table \ref{tab:results-test}}.

The bias between low-recall and high-precision values indicates more false negatives. To choose the right model and reduce the bias between precision and recall in the results, the beta value for the F measure has been set to 1.2 (as evaluated using \textbf{Table \ref{tab:f1analysis}}), which slightly reduces the bias towards precision. This score is used to conduct A/B testing to select the best model.

The near-identical performance of the model in the training set and the testing set shows that the model has been trained well and generalized correctly, given that the data in the training and test set follow a similar distribution.

\subsection{LSTM Model Performance}
After running multiple simulations, it was concluded that the best LSTM model that helped predict codes with a promising accuracy of 97\% was the one used after up-sampling. The base model (without up-sampling) performs with an accuracy of 92\%. Before up-sampling, 19 out of 23 codes in the test set performed with precision above 90\%; however, after up-sampling, the number increased to 20.

\subsection{DNN Model Performance}
DNN, when trained with the original data, achieves an overall accuracy of $88\%$ for training and $83\%$ for the test set.  After using the up-sampled data, the model has attained a remarkable accuracy of $95\%$.Before up-sampling, 15 out of 23 codes in the test set were performed with precision above 90\%; however, after up-sampling, the number increased to 19. 

\subsection{Text-CNN Model Performance}
The overall accuracy achieved by Text-CNN is 98\%. 19 out of 23 harmonized codes were correctly classified with the base data and 20 out of 23 with the up-sampled data. We considered the threshold of above 90 percent to determine that the code is correctly classified.

\subsection{Performance Evaluation}
Based on our evaluation metrics (accuracy, precision, and recall) for each code and the respective models, we can say that Text-CNN gave the best results.

As presented in \textbf{Table \ref{tab:results-test}} by comparing the number of codes performed with low (below 0.80), medium (between 0.80- 0.90) or high (above 0.90) range of precision, it can be seen that the Text-CNN approach has 19 codes in high range for the model before up-sampling. After up-sampling, this count increased to 20 codes. Similarly, 20 codes performed with high recall before up-sampling for Text-CNN. After up-sampling, the count increased to 21. These counts are higher than those of the LSTM or DNN approach. In addition, Text-CNN has a high accuracy of 98\%, making it the best-performing model.

\section{MLOps}
MLOps helps to manage the code base, handle collaboration between developers, and integrate between environments with release management and version control. With the help of these practices, it is possible to automate the deployment of ML models in a large-scale production environment without the overhead of model monitoring and evaluation while creating a reproducible workflow. Keeping these benefits in mind, a machine learning life cycle with continuous integration, training, deployment, and monitoring has been implemented \cite{b21}, \cite{b22}, \cite{b23}.

\subsection{Infrastructure}

\begin{table}[ht]
\centering
\begin{tabular}{ | l | c | c | }
\hline
\textbf{Hyper-parameters} & \textbf{Range} & \textbf{Values}  \\ \hline
\textbf{Filter Size}  & [64,128,256]        & 128                                       \\ \hline
\textbf{Kernel Size}       & 1 - 10     & [5]                 \\ \hline
\textbf{Embedding Dim}    & 50  - 150   &   [100]              \\ \hline
\textbf{N\_layer}         & 1.0 - 5.0     & 1.0                 \\ \hline
\end{tabular}
\caption{Hyper-parameters for Text-CNN Model}
\label{tab:hyp-TextCNN}
\end{table}

\begin{table*}[hbt!]
\centering
\begin{tabular}{ | c | c | c | c |}
\hline
\textbf{Precision} & \textbf{Recall} & \textbf{Evaluation}  & \textbf{Type of F Score} 
\\ \hline
High & 
High & 
\shortstack{Best case scenario - model detects HS code and predicts well with less misclassifications \\ performs well on train and test} & 
F0  
\\ \hline
High &
Low & 
\shortstack {Model detects HS codes but with some misclassifications  model performs better on training \\ but not that well on tests}  &
between F1-F2
\\ \hline
Low & 
High & 
Over fitting -model predicts well on train but not on test set & 
between F0-F1  
\\ \hline
Low & 
Low & 
Model is not learning as Precision and recall are low in both train and test & 
between F1-F2 
\\ \hline
\end{tabular}
\caption{F Score estimation to balance bias between Precision and Recall}
\label{tab:f1analysis}
\end{table*}

\begin{table*}[t]
\centering
\begin{tabular}{|l|l|l|l|l|l|l|}
\hline
\textbf{Model}   & \multicolumn{2}{c|}{\textbf{\begin{tabular}[c]{@{}c@{}}Hyper-parameter Tuning\end{tabular}}} & \multicolumn{2}{c|}{\textbf{Training}} & \multicolumn{2}{c|}{\textbf{Total Time Taken}} \\ \hline
                 & \textbf{Base Model}                                & \textbf{up-sampled Model}                               & \textbf{Base Model}   & \textbf{up-sampled Model}   & \textbf{Base Model}       & \textbf{up-sampled Model}       \\ \hline
\textbf{DNN}     & 36 mins                                      & 2 hours 55 mins                                  & 15 mins         & 27 mins              & 51 mins             & 3 hours 36 mins          \\ \hline
\textbf{LSTM}    &     4 hours                                         &     4.2 hours                                             &       1.5 hours          &     1.2 hours                 &    5.5 hours                 &         5.4 hours                 \\ \hline
\textbf{Text-CNN} &     2 hours 20 min                                        &   1 hour 50 mins                                               &         55 mins        &             1 hour 20 mins        &              3 hours 15 mins       &       3 hours 10 mins                   \\ \hline
\end{tabular}
\caption{Time taken for training proposed models}
\label{tab:time-profiling}
\end{table*}
\subsubsection{CI-CD}
Continuous Integration(CI) in ML means that the machine pipeline will automatically update each time we update the code/data. Continuous deployment (CD) releases the new changes to the customer after passing each stage of the production pipeline. CI-CD helps with remote deployment to production and significantly reduces the time taken for integration and deployment.

\subsubsection{Tools and Services}
AWS offers a wide variety of tools and services for MLOps and CI-CD implementation. It is up to the user to select the right set of services and tools. After following a methodical approach to system design and planning, the following tools and services for implementation have been selected: 
\begin{itemize}
    \item {\verb|AWS S3:|} An object storage platform. In addition to object storage, a file storage system such as a relational database can also be considered. To store the data used for training, along with model weights and configuration files in well organized folders
    \item {\verb|GitHub Enterprise:|} At an enterprise level, source code management encompasses the concurrent sharing, storage, and collaborative development of source code across distinct repositories dedicated to tasks such as data validation, pre-processing, training, and inference.
    \item {\verb|AWS CodeBuild:|} It is a fully managed build service that helps to create and run a docker image. In addition, it automates the compilation and quality assessment of the code and eliminates the need to manage the underlying infrastructure.
    \item {\verb|AWS ECR:|} Elastic Container Registry to store the docker image and helps to manage versioning of docker images.
    \item {\verb|AWS EventBridge:|} A serverless event trigger for predefined schedules, such as code changes or changes in source data. It helps in tracking events and trigger workflows.
    \item {\verb|AWS Step Functions:|} A service that builds and maintains workflow orchestration.
    \item {\verb|AWS State Machine:|} A compute resource, such as a server, consists of a collection of states and the relationships among those states. There are several types of compute resources available as state machines, such as glue,lambda, and EC2.
    \item {\verb|AWS SageMaker:|} It is a fully managed compute API service that provides the ability to quickly build, train, and deploy ML models. Jupyter notebook instances were used to build the code. The instance ml.p2.8xlarge has been used to train and run our models.
\end{itemize}

\subsubsection{Serverless Architecture}
Serverless architecture is an approach that allows developers to build and run models without managing the underlying infrastructure; in this case, compute instances. In traditional methods, teams must maintain server hardware, take care of software and security updates, create backups in case of failure, and manually scale compute resources depending upon demand. By adopting a serverless architecture, developers can off-load these responsibilities to the cloud provider, enabling them to focus on writing application code.

\begin{figure}[h]
    \centering
    \includegraphics[width=1\linewidth]{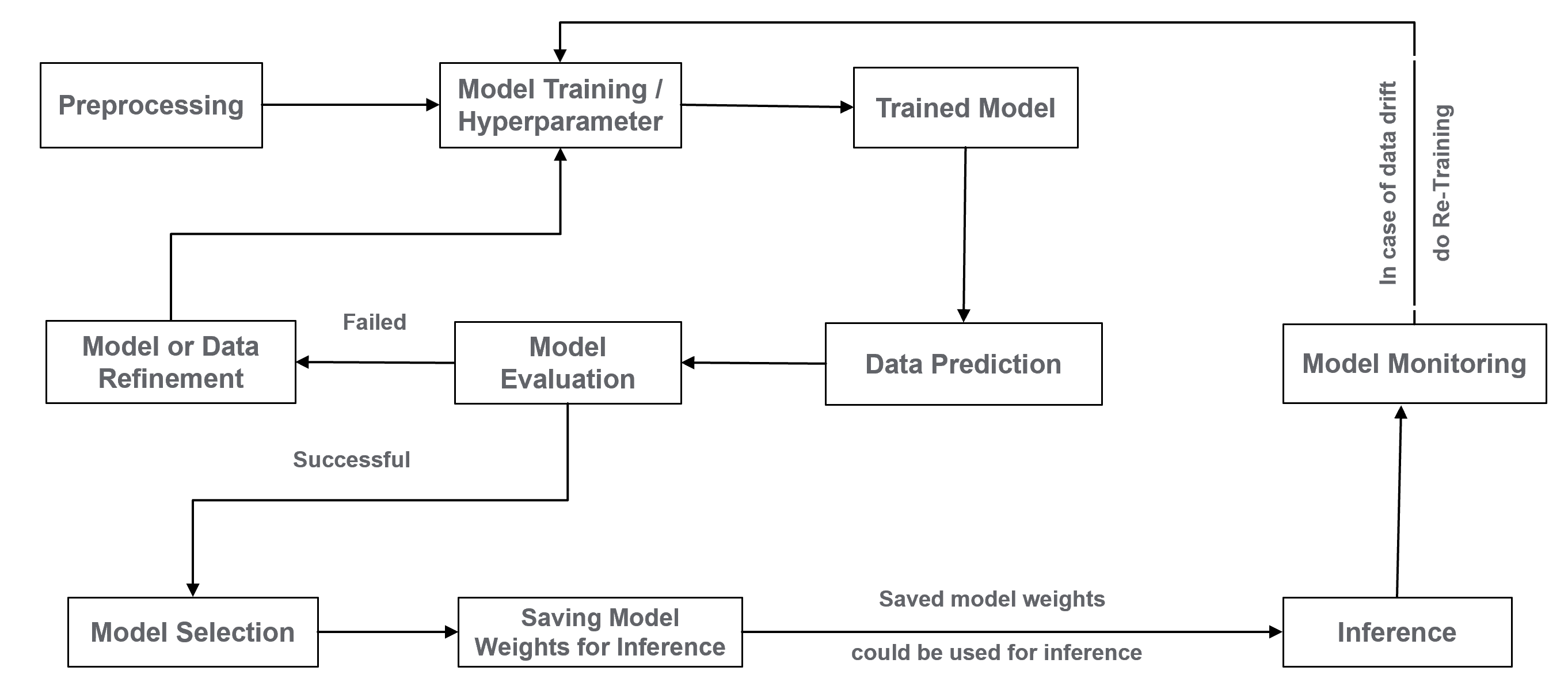}
    \caption{Machine learning model life-cycle}
    \label{fig:Modellifecycle}
\end{figure}

\begin{figure}[h]
    \centering
    \includegraphics[width=1\linewidth]{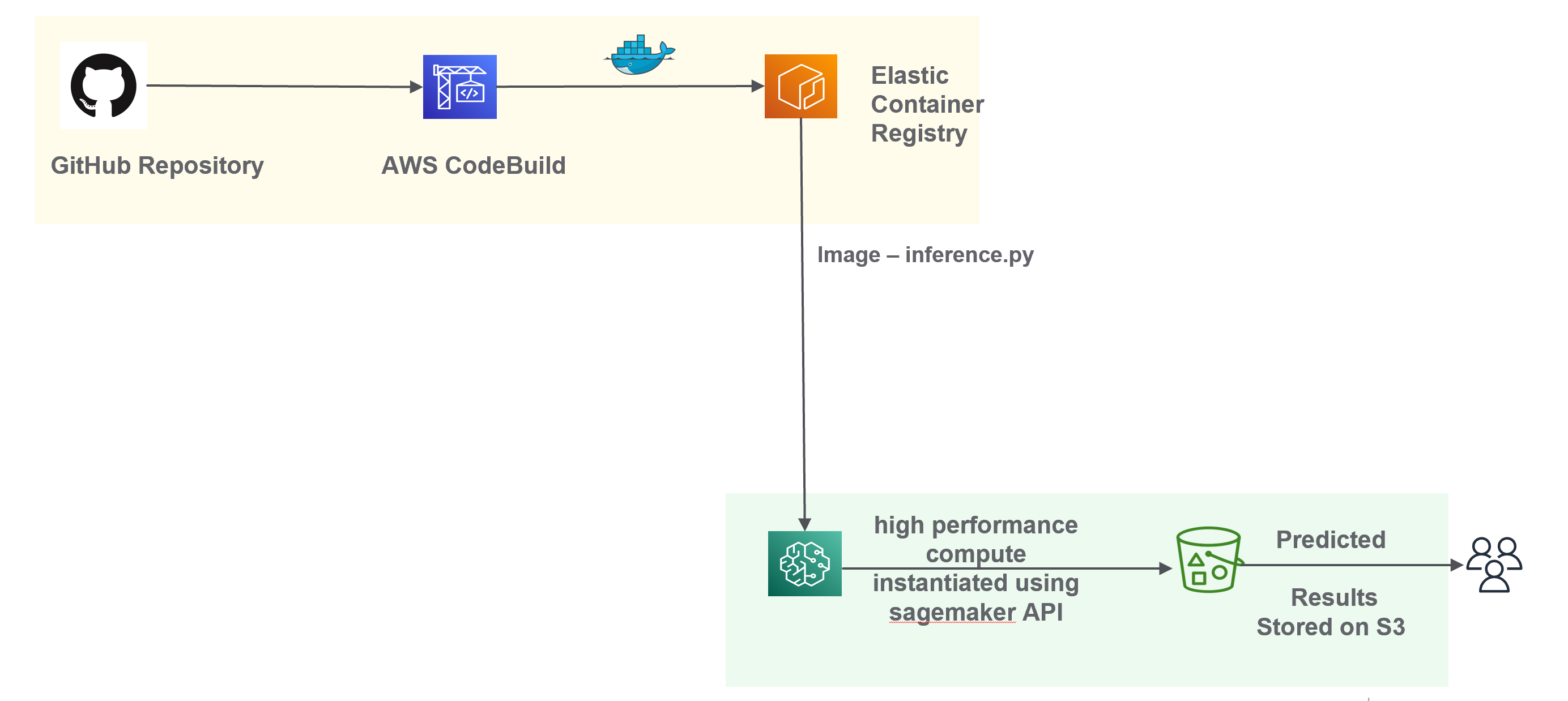}
    \caption{Serverless Architecture}
    \label{fig:serverless}
\end{figure}

Developers can write and deploy code, while a cloud provider provides servers to run their models at any scale. As shown in \textbf{Figure \ref{fig:serverless}}, the application code is residing in a GitHub repository. This repository is linked to AWS CodeBuild, which creates the Docker image of the code and pushes the image to the Elastic Container Registry. From a low-power instance (here, SageMaker as a compute), any compute instance as per requirements can be called through an API. These compute instances are used on a per-need basis, i.e., they automatically scale up when there are jobs to run and scale down when there are no jobs left in the queue. This architecture provides the flexibility to offload the management of any compute instances to implement the ML model lifecycle. A sample architecture diagram is shown in \textbf{Figure \ref{fig:Modellifecycle}}.

\subsection{ML Lifecycle}
The machine learning life cycle is a process that involves building data pipelines in phases (microservice architecture approach). As seen in \textbf{Figure \ref{fig:Inference}}, two pipelines have been built, one for inference and one for retraining. Each pipeline has detailed stages to implement the end-to-end life cycle successfully. The critical stages for the ML life cycle are as follows.

\begin{itemize}
\item {\verb|Data collection and pre processing:|} The data can come from many sources. As listed in \ref{DataPreparation}, the preprocessing stages should be well defined, and it is beneficial to collect a good representation of data points across the population during the learning stage to develop a well-defined preprocessing pipeline.
\item {\verb|Model Training:|} In model training, a part of the training data is used to find coefficients of weight (model parameters) for the models. This process helps minimize the error for the given data set. The model is trained with different hyperparameters and the best configurations are selected by applying the model to the validation data. The remaining data are then used to test the model.
\item {\verb|Obtain Model weights:|} The model is trained on the entire training data set to obtain the final model weights.
\item {\verb|Evaluation of the trained model:|} Test data is used to evaluate the weights of the trained model using various performance metrics such as accuracy, precision, and recall. 
\item {\verb|Model selection:|} Model selection is performed using AB testing to obtain the significance of the results for the evaluated models. With AB testing, the best performing model is selected using evaluation metrics. The respective model weights are saved for the inference pipeline.
\item {\verb|Inference:|} In the Inference stage, the saved model weights are used to make predictions on the test data in real time.
\item {\verb|Monitoring:|} The distribution and results are continuously monitored. Any significant drift in the data sets can trigger the retraining pipeline.
\end{itemize}

\subsubsection{Model Inference}
\label{ModelInference}
\begin{figure}[h]
    \centering
    \includegraphics[width=1\linewidth]{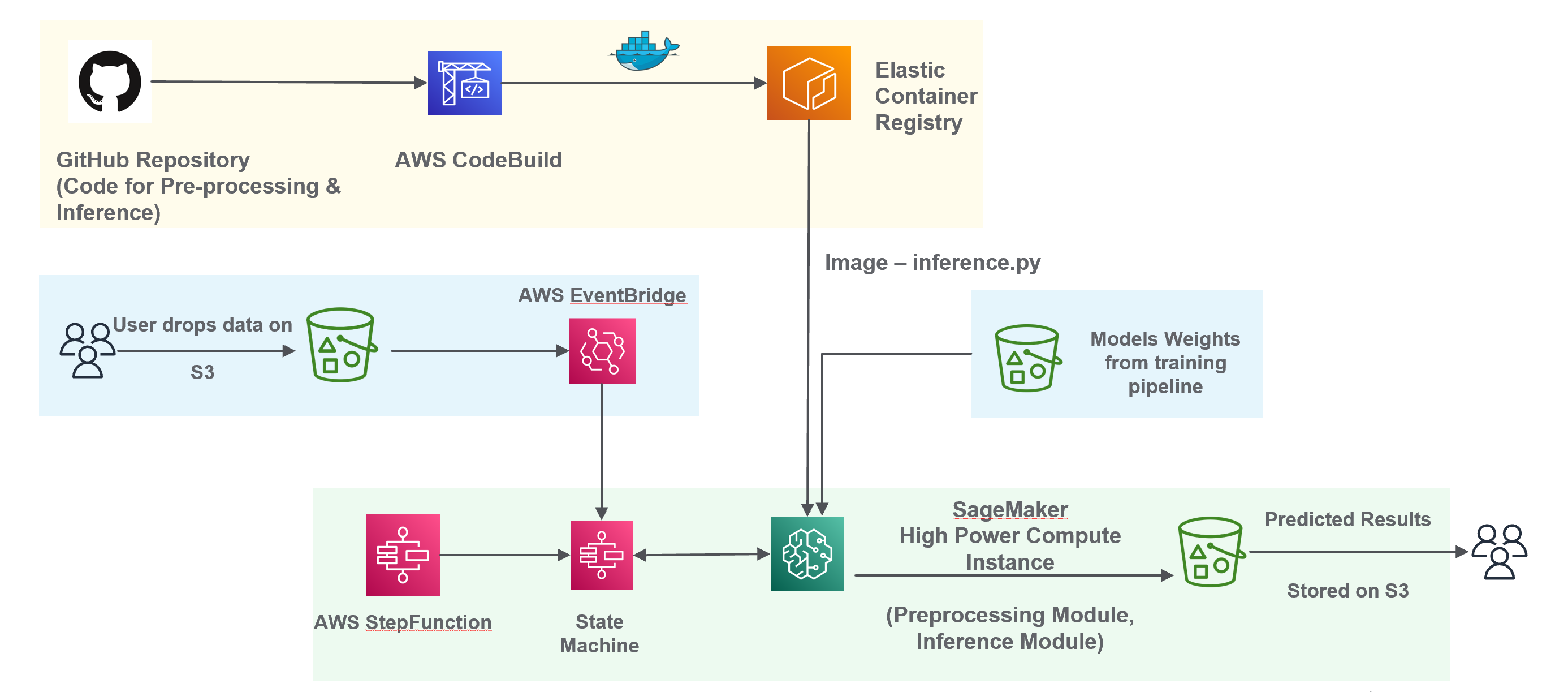}
    \caption{Inference pipeline}
    \label{fig:Inference}
\end{figure}

The main objective of the inference pipeline is to provide a seamless process for generating predictions for the user in real-time. The pipeline should start automatically every time a user provides data for predictions(predefined event) either using object storage or through the API. The predicted results of the model should be returned to the user as shown in \textbf{Figure \ref{fig:Inference}}.

Steps in the Inference pipeline:
\begin{itemize}
\item Connect the GitHub code repository with a building project in AWS CodeBuild to create docker images. This connection between GitHub and CodeBuild includes a web-hook that will automatically trigger the project to run and build a new docker image whenever there is a push to the master branch of the GitHub repository
\item Once the image is built, CodeBuild automatically pushes the updated docker image to ECR.
\item To automate the workflow of the model life cycle, step functions are used to create State Machine ARNs. State machines can be created using the AWS Sagemaker and Step Function SDK.
\item In the State Machine, the workflow is designed by organizing steps of the ML pipeline in sequential order (pre-processing followed by inference). 
\item The State Machine is triggered using an event trigger service, i.e., AWS EventBridge, where rules to start the State Machine workflow on predefined events (whenever the user drops new test data on S3 for batch inference) can be created.
\end{itemize}

\subsubsection{Model Retraining}
The main objective of the retraining pipeline is to trigger the training process when the user drops data for retraining after evaluation and corrections or when the model performance starts to deteriorate (data drift) \cite{b25} ,\cite{b24}.
The architecture of the retraining pipeline follows an approach similar to that of the inference pipeline. 

\begin{figure}[h]
    \centering
    \includegraphics[width=1\linewidth]{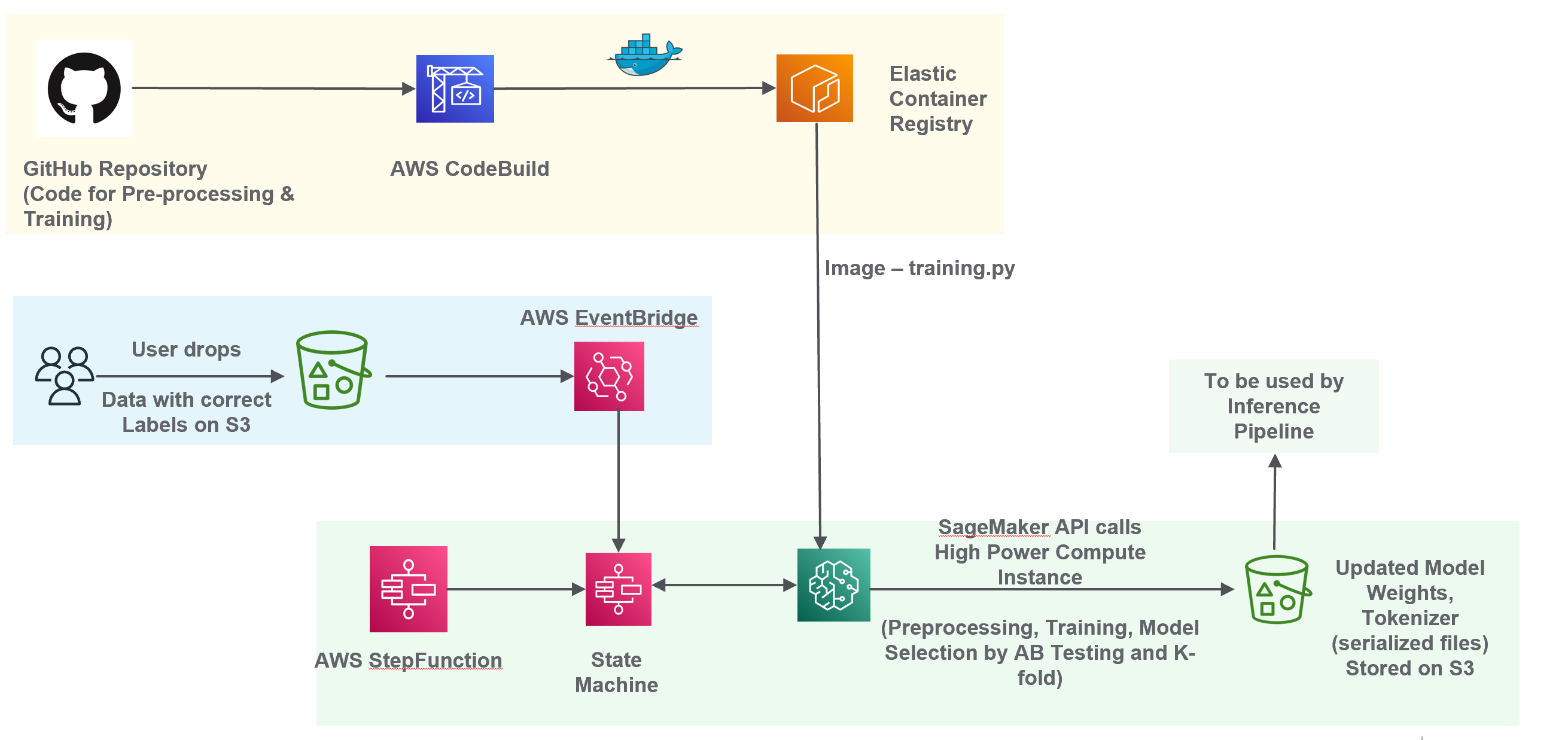}
    \caption{Retraining pipeline}
    \label{fig:Retraining}
\end{figure}

\subsection{Implementation Challenges}
\begin{itemize}
\item {\textbf{Interoperability between services:}} There is always a challenge in making services talk to each other through network security. It is not easy to whitelist services between AWS accounts when working on a serverless architecture. Services are segregated between different AWS accounts for data privacy and security reasons.
\item {\textbf{Identifying right services:}}
AWS provides a lot of flexibility in terms of the services that can be utilized for serverless architecture ranging from lambda to custom instances using Sagemaker. Therefore, finding the balance between industrialization, operationalization, cost, maintainability, and scalability was a key challenge.
\end{itemize}
\section{Conclusion}
\subsection{Risks Involved}
\subsubsection{Customs Code Alignment}
Each country differs in their criteria for assigning HS codes. This leads to multiple codes assigned for the same product in different regions. When models train on such data, they can be confused if there are multiple codes for the same set of descriptions.
\subsubsection{Changes in HS Code Descriptions}
Based on our analysis, the HS code descriptions change from time to time. Due to this, the model cannot learn certain HS codes with high efficiency.
\subsubsection{Class Imbalance}
Of the existing HS codes, some of them have a representation less than 4\% throughout the training data set. This results in class imbalance, and the data distribution should be monitored for any bias while retraining due to imbalance.
\subsubsection{Test split on Data}
In the training phase to validate the model, the data was split into 95\% for training and 5\% for testing to train the model on more data. In the future, as data accumulate for each HS code and the retraining stages evolve, the split of 5\% will be increased to better view the model performance in the test data set. 
\subsubsection{Updates to HS Code by World Customs Org. (WCO)}
WCO updates the HS codes every 5 to 6 years. These updates could affect model predictions if the model is not re-trained with updated HS codes.
\subsubsection{Introduction of new products}
When new products are introduced into the data set, an HS code is required. It will not be possible for a model to predict a code accurately if it has not previously trained on the new product. Such new products will require manual assignments until enough samples are available to retrain the ML models.
\subsection{Future Scope}
The limitations of this analytical implementation are directed toward the topics that can be addressed in the future. The following are some areas of improvement.
\begin{itemize}
\item Use additional categorical features from the given dataset to improve model predictions
\item Implement alternate up-sampling techniques such as SMOTE
\item Expand the models for other regions apart from Europe and North America
\item Include model monitoring strategies like KL Divergence can help to overcome data drift while increasing model performance
\item Incorporate parallel processing to reduce training time further
\item Implement cloud formation template for automated code deployment between environments
\item Future work will also include the potential integration of generative AI approaches in place of current architectures. This transition will be considered once generative models exhibit greater transparency and when institutional data governance policies allow the secure sharing of confidential data for model training and inference. 
\end{itemize}


\end{document}